\documentclass[a4paper,twoside]{article}

\usepackage{epsfig}
\usepackage{subcaption}
\usepackage{calc}
\usepackage{amssymb}
\usepackage{amstext}
\usepackage{amsmath}
\usepackage{amsthm}
\usepackage{multicol}
\usepackage{pslatex}
\usepackage{apalike}
\usepackage{algorithm2e}
\usepackage[bottom]{footmisc}

\usepackage{hyperref}

\usepackage{SCITEPRESS}

\usepackage[export]{adjustbox}
\usepackage{mathtools}

\newcommand{\Ie}{I.e.\@}
\newcommand{\ie}{i.e.\@}

\newcommand{\eg}{e.g.\@}

\newcommand{\ia}{i.a.\@}

\newcommand{\resulturl}{\url{https://github.com/dasec/dataset-duplicates}}

\newcommand{\customsectiontext}[1]{\uppercase{#1}}

\begin{document}

\title{Double Trouble?\\ Impact and Detection of Duplicates in Face Image Datasets}
\author{\authorname{Torsten Schlett\sup{1}\orcidAuthor{0000-0003-0052-2741},
Christian Rathgeb\sup{1}\orcidAuthor{0000-0003-1901-9468},
Juan Tapia\sup{1}\orcidAuthor{0000-0001-9159-4075} and
Christoph Busch\sup{1}\orcidAuthor{0000-0002-9159-2923}}
\affiliation{\sup{1}da/sec - Biometrics and Security Research Group, Hochschule Darmstadt, Germany}
\email{\{torsten.schlett, christian.rathgeb, juan.tapia-farias, christoph.busch\}@h-da.de}
}


\abstract{Various face image datasets intended for facial biometrics research were created via web-scraping, \ie{} the collection of images publicly available on the internet.
This work presents an approach to detect both exactly and nearly identical face image duplicates, using file and image hashes.
The approach is extended through the use of face image preprocessing.
Additional steps based on face recognition and face image quality assessment models reduce false positives,
and facilitate the deduplication of the face images both for intra- and inter-subject duplicate sets.
The presented approach is applied to five datasets, namely LFW, TinyFace, Adience, CASIA-WebFace, and C-MS-Celeb (a cleaned MS-Celeb-1M variant).
Duplicates are detected within every dataset, with hundreds to hundreds of thousands of duplicates for all except LFW.
Face recognition and quality assessment experiments indicate a minor impact on the results through the duplicate removal.
The final deduplication data is made available at \resulturl{}.}

\onecolumn \maketitle \normalsize \setcounter{footnote}{0} \vfill

\section{\customsectiontext{Introduction}}
\vspace{-0.5em}

Face recognition or other facial biometrics research often involves web-scraped face image datasets, \eg{} to train or evaluate face recognition models.
Scraping face images from the web can accidentally lead to the inclusion of mislabelled or duplicated images.
This paper presents a duplicate detection approach that searches for both exact and near duplicates,
which is applied to a selection of five web-scraped face image datasets:
LFW \cite{LFWTech},
TinyFace \cite{Cheng-TinyFace-LowResolutionFaceRecognition-ACCV-2018},
Adience (aligned) \cite{Eidinger-AgeGenderEstimationUnfilteredFaces-TIFS-2014},
CASIA-WebFace \cite{Yi-LearningFaceRepresentationFromScratchCASIAWebFace-arXiv-2014},
and C-MS-Celeb (aligned) \cite{Jin-CMSCeleb-2018}.
Additional steps are presented to mitigate false positives and to systematically deduplicate the datasets through the use of face recognition and face image quality assessment models.
The final deduplication data is made publicly available\footnote{\resulturl{}},
and the general methodology should be applicable to any typical web-scraped face image datasets beyond the ones examined in this paper.
The effect of the duplicate removal on face recognition and face image quality assessment experiments is examined as well.

The rest of this paper is structured as follows:

\begin{itemize}
\item Related work is discussed in \autoref{sec:related-work}.
\item The fundamental duplicate detection approach is presented in \autoref{sec:duplicate-detection}.
\item Preservative deduplication steps after the initial duplicate detection are described in \autoref{sec:deduplication}.
\item Effects of duplicate removal on face recognition and face image quality assessment are investigated in \autoref{sec:effects}.
\item Findings are summarized in \autoref{sec:summary}.
\end{itemize}

\vspace{-1em}
\section{\customsectiontext{Related work}}
\label{sec:related-work}

To the best of our knowledge there is no closely related facial biometrics work that proposes a similar face image duplicate detection based on file and image hashes (\autoref{sec:duplicate-detection}), with subsequent preservative deduplication based on face recognition and face image quality assessment (\autoref{sec:deduplication}), including an examination of multiple existing separate web-scraped face image datasets.
There are however various more loosely related works on automated face image dataset label cleaning:

In \cite{Jin-CMSCeleb-2018} a graph-based label cleaning method is presented and applied to the MS-Celeb-1M \cite{Guo-Face-MSCeleb1M-ECCV-2016} dataset, resulting in the C-MS-Celeb dataset that is further examined in this work.
This approach first extracts face image feature vectors using a pretrained model,
based on which similarity graphs are constructed between the images of each original subject label.
The cleanup then consists of the deletion of insufficiently similar graph edges and the application of a graph community detection algorithm.
Only sufficiently large communities are retained.
An additional step computes the similarity of remaining images to the retained community feature centers,
which may lead to the assignment of these images to a sufficiently similar community.

This work in contrast aims to detect face image duplicates across the examined datasets using general file and image hashes, without any initial reliance on subject labels or facial feature vectors. The latter are however used in an additional but technically optional preservative deduplication phase (in addition to face image quality assessment).
As this work examines C-MS-Celeb itself, it is shown that the original MS-Celeb-1M \cite{Guo-Face-MSCeleb1M-ECCV-2016} collection and the C-MS-Celeb cleaning approach \cite{Jin-CMSCeleb-2018} did not suffice to remove a substantial number of apparent duplicates (over 500,000 are removed at the end of this work in \autoref{sec:deduplication}).

Another cleaning approach applied to \ia{} MS-Celeb-1M \cite{Guo-Face-MSCeleb1M-ECCV-2016} is presented in \cite{Jazaery-FaceCleaningQuality-IETBiometrics-2019},
albeit without a comparison to \cite{Jin-CMSCeleb-2018}.
This other approach first selects a ``reference set'' of images for each subject, based on face image quality assessment (and on web search engine rankings for the images, if available).
More specifically the images with quality scores above the mean quality score are selected within the subject sets.
The following cleaning steps then either keep or discard the other images of each subject's set,
based on whether each image's mean similarity score to the ``reference set'' images is above a threshold that varies depending on the image quality score and parameters configured by the researchers (through the examination of a data subset).

This is again in contrast to this work's duplicate detection objective, which this label cleaning doesn't conceptually address.
A relation is the use of both quality assessment and similarity scores.
But in this work, besides the use of different models, these are used in the aforementioned preservative deduplication phase, which builds upon the independent file/image hash duplicate detection.
The way in which the quality and similarity scores are considered differs as well,
and \eg{} includes the resolution of inter-subject duplicates (instead of only filtering within each isolated image set per subject).

In \cite{Zhang-CelebritiesOnTheWeb-TMM-2012} the ``Celebrities on the Web'' dataset was constructed, utilizing names found in textual data collected alongside the images to establish initial subject label candidates, with the final subject label assignment being facilitated through the use of a face image similarity graph based on the initial labels.
This is more distantly related work as it pertains to the collection of dataset images from scratch, instead of duplicate detection or label cleaning for existing face image datasets, which may not typically provide the textual data that is used by this approach.
Nevertheless, part of this approach involves the intentional web search for near duplicate images with the help of image hashing, to gather further text data.
The approach can also handle multiple faces in collected images.

In this work the base duplicate detection in contrast only requires images (without subject labels or text), which likewise do not necessarily have to be images that show a single face.
The additional preservative deduplication however expects existing subject labels and assumes that each image is supposed to show exactly one face,
both of which should be the case for typical already constructed face image datasets.

\section{\customsectiontext{Duplicate Detection}}
\label{sec:duplicate-detection}

This section describes the fundamental duplicate detection approach, which considers exact and near duplicates.

\subsection{Exact Duplicates}

Exact duplicates can technically be found purely via exact data comparisons.
For the sake of computational efficiency we use BLAKE3 \cite{BLAKE3} hashes of the file data to collect the initial duplicate sets.
File hash false negative duplicates are impossible, since the same data always results in the same hash value,
but false positives are technically possible (hash collisions).
Therefore an additional step which checks for fully identical file data is employed within each duplicate set found by the file hash step, to ensure that there are no false positives.

Although this kind of exact duplicate check is arguably simple to conduct,
exact duplicates were nevertheless found in all of the examined datasets,
implying that no such check was performed as part of the datasets' creation.

\subsection{Near Duplicates}
\label{sec:near-duplicates}

There can also be images that differ slightly,
yet are arguably so similar that they should be counted as near duplicates for typical facial biometrics research.

\begin{figure}
\centering
\begin{tabular}{c|c|c}
\includegraphics[height=2.2cm]{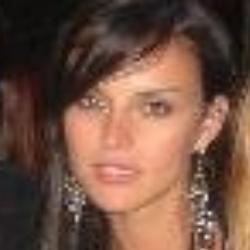} &
\includegraphics[height=2.2cm]{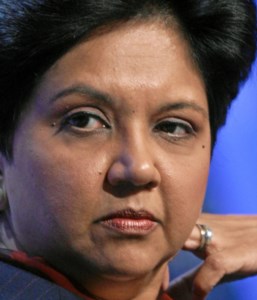} &
\includegraphics[height=2.2cm]{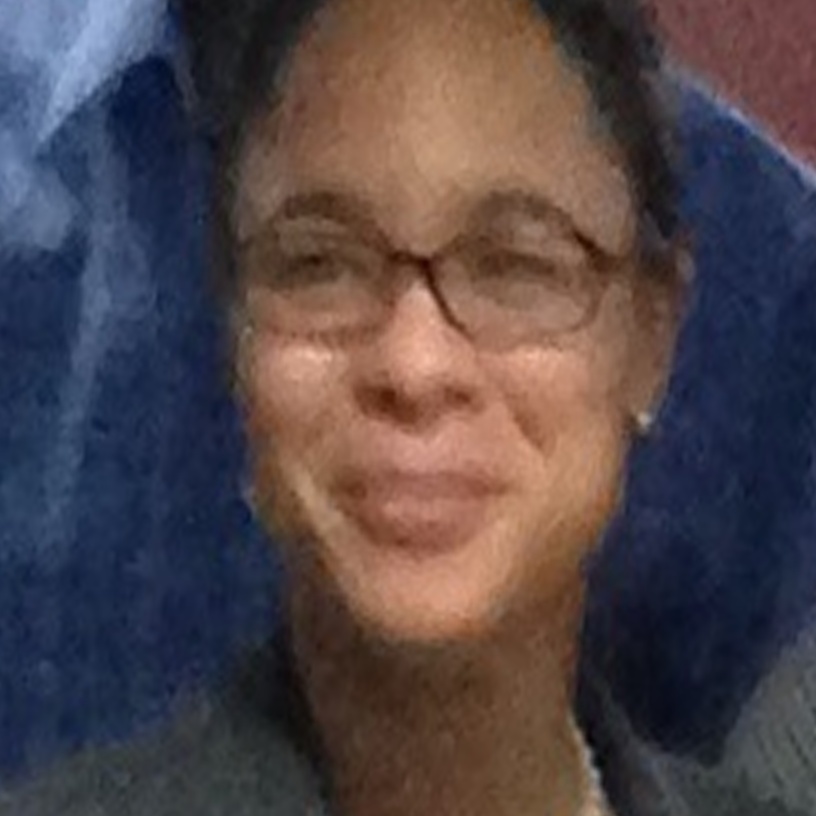}
\\
\includegraphics[height=2.2cm]{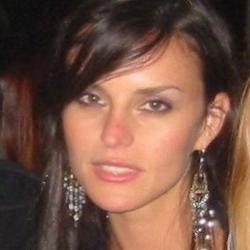} &
\includegraphics[height=2.2cm]{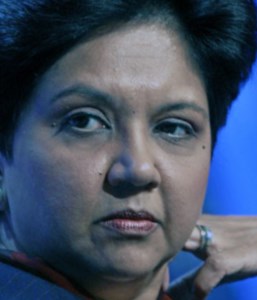} &
\includegraphics[height=2.2cm]{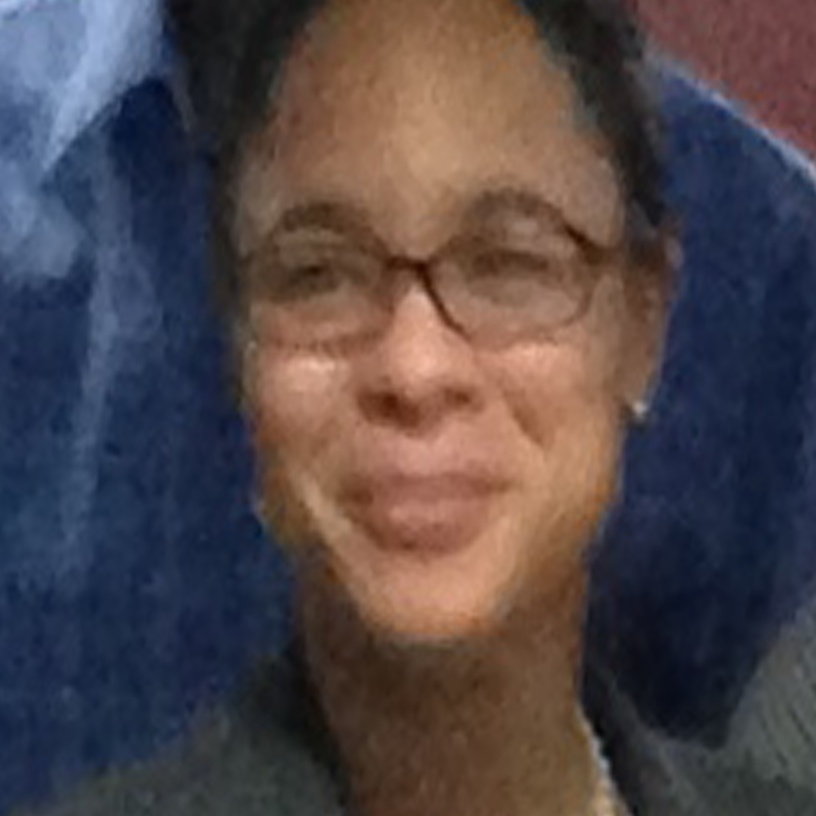}
\end{tabular}
\caption{\label{fig:example-duplicates} Three near duplicate pair examples, all detected by pHash. From left to right, the first example is from CASIA-WebFace, the second from C-MS-Celeb, and the third from Adience.}
\end{figure}

While it is well-defined what exact duplicates are,
as files are either exactly identical or not,
there are many different possible ways in which near duplicates can be defined.
In this paper we employ two image hash implementations with default settings from the ``ImageHash'' Python package at version 4.3.1\footnote{\url{https://pypi.org/project/ImageHash/}},
based on cursory manual examinations of found duplicates prior to the main experiments:
``pHash'' (for ``perceptual hashing'') and ``crop-resistant hashing''.
See \autoref{fig:example-duplicates} for near duplicate examples.

The pHash method resulted in more detected duplicates across all examined datasets than the crop-resistant hashing method,
and so the latter may be comparatively less suitable for this particular use case (\ie{} at least with the used default configuration).

As the duplicate sets found by different image hash functions can overlap, and since image hashes may also lead to false positive duplicates, additional set merging and false positive correction should be employed as part of the final preservative deduplication (\ie{} deduplication that aims to keep one image per duplicate set, instead of simply removing all possible duplicates).
This is further described in \autoref{sec:false-positive-correction}.

\subsection{Preprocessed Face Images}
\label{sec:preprocessed-duplicates}

\begin{figure}
\centering
\newcommand{\incg}[1]{\includegraphics[height=2cm,valign=c]{img/fdet/#1}}
\begin{tabular}{ccc}
Original & Face detection & Preprocessed \\
\incg{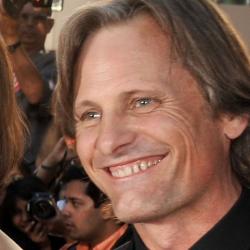} & \incg{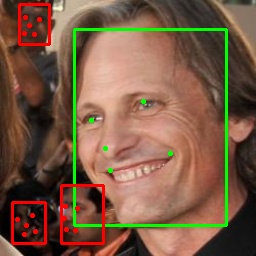} & \incg{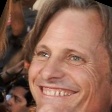}
\end{tabular}\\
\caption{\label{fig:preprocessing} Preprocessing example using an image from the CASIA-WebFace dataset.}
\end{figure}

Various face recognition and face image quality assessment models rely on face image preprocessing,
which typically crops and aligns the original face image based on detected facial landmarks.
The just described duplicate detection can optionally be carried out using such preprocessed face images.
This should be done as an additional step after the duplicate detection on the unmodified original images,
as the facial landmark detection required for preprocessing may fail for some images,
and because the preprocessed/original image variants can both yield duplicate sets that the other image variant does not.

In this paper the similarity transformation as employed for ArcFace \cite{Deng-ArcFace-IEEE-CVPR-2019} is used for face image preprocessing.
The SCRFD-10GF model buffalo\_l from InsightFace \cite{Guo-SCRFD-ICLR-2022}\footnote{\url{https://github.com/deepinsight/insightface/tree/master/python-package}}
is used to obtain the facial landmarks that are required by this preprocessing step.
For images in which multiple faces are detected, a primary face is selected based on the detection's bounding box width and height, its proximity to the image center, and the detector confidence score.
The preprocessed images all have the same 112×112 width and height.
\autoref{fig:preprocessing} shows a preprocessing example.

Face images for which the landmark detection failed amount to
0 for LFW,
859 for TinyFace (only 4 of which are duplicates based on the detection on the original images),
79 for Adience (9 of which are duplicates),
129 for CASIA-WebFace (2 of which are duplicates),
and 6,179 for C-MS-Celeb (351 of which are duplicates).
These images are simply not considered by this additional duplicate detection step.

\subsection{Examined Dataset Duplicates}
\label{sec:datasets}

\autoref{tab:dataset-overview} shows an overview of the examined datasets, including the total image counts and the duplicate counts.
Intra-subject duplicates are duplicates found within the same subject.
\Ie{} the identical or very similar images belong to the same subject (are stored as mated samples), and are presumed to stem from different capture attempts or sessions, but in fact stem from the very same capture attempt.
Inter-subject duplicates on the other hand are found across different subjects.

LFW, TinyFace, and Adience each have under 20,000 face images.
CASIA-WebFace is substantially larger with 494,414 face images,
which is however still closer to the three smaller datasets than to C-MS-Celeb with its 6,464,016 face images.
The number of different subjects does not increase as sharply as the number of total face images for the larger datasets.

Among the datasets, LFW is the only one with a very low number of duplicates.
The evaluation in \autoref{sec:effects}, which will investigate the impact on facial biometrics, consequently omits the LFW dataset.

The TinyFace dataset includes 153,428 non-face images, which are not examined in this paper.
Only the remaining 15,975 face images are considered.

For the Adience dataset we more specifically use the ``aligned'' version of the face images.

Although the absolute number of duplicates for CASIA-WebFace  is higher than for the smaller datasets,
the duplicate percentage with respect to the face image total is relatively low (excluding LFW).

The C-MS-Celeb dataset is a subset of MS-Celeb-1M \cite{Guo-Face-MSCeleb1M-ECCV-2016} with cleaned subject labels.
In this paper the ``aligned'' image variants are used.
Both the absolute number of duplicates as well as the duplicate percentage with respect to the face image total is the highest among all examined datasets.
For this dataset 33,918 inter-subject duplicates simultaneously are intra-subject duplicates.
The combined count of images that are part of some duplicate set is thus 885,476.
This intra-/inter-subject duplicate overlap did not occur for the other datasets,
so the combined count for each of these simply is the sum of the intra- and inter-subject counts.

\begin{table*}[h]
\centering
\caption{\label{tab:dataset-overview}Image, subject, and duplicate counts for the examined datasets prior to the preservative deduplication in \autoref{sec:deduplication}. \\
Intra: Intra-subject duplicates (all images in duplicate sets belonging to only one subject). \\
Subjects-w.-intra: Subjects with at least one intra-subject duplicate. \\
Inter: Inter-subject duplicates (all images in duplicate sets belonging to multiple subjects). \\
Subjects-w.-inter: Subjects with at least one inter-subject duplicate.
}
\begin{tabular}{|r|r|r|r|r|r|r|}
\hline
      Dataset &    Images & Subjects &   Intra & Subjects-w.-intra &   Inter & Subjects-w.-inter \\
\hline
          LFW &    13,233 &    5,749 &       6 &                 3 &       6 &                 6 \\
     TinyFace &    15,975 &    5,139 &     662 &               286 &      53 &                52 \\
      Adience &    19,370 &    2,284 &   1,609 &               274 &       4 &                 4 \\
CASIA-WebFace &   494,414 &   10,575 &   9,614 &             2,677 &     288 &               186 \\
   C-MS-Celeb & 6,464,016 &   94,682 & 753,277 &            76,841 & 166,117 &            27,895 \\
\hline
\end{tabular}
\end{table*}

Besides duplicate detection within each dataset, the approach can also be applied to check for overlap between the datasets.
No such inter-dataset duplicate cases were detected using the original images.
Some cases were however found using the preprocessed images,
and a subset has been manually confirmed to be true positives (predominantly but not exclusively between CASIA-WebFace and C-MS-Celeb).
Although a more extensive dataset overlap investigation is outside the scope of this work,
this does indicate that the approach of this paper could be used as a part of future work on this topic.

\vspace{-1em}
\section{\customsectiontext{Preservative Deduplication}}
\label{sec:deduplication}

This section describes additional preservative deduplication steps that continue from the fundamental duplicate detection of the prior section.
These steps keep one selected image per duplicate set (\autoref{sec:exact-intra-deduplication}, \autoref{sec:quality-based-deduplication}) and mitigate false positive duplicate detections (\autoref{sec:false-positive-correction}).
They also move the kept image of an inter-subject duplicate set to the most fitting subject within that set, based on similarity scores (\autoref{sec:inter-subject-deduplication}).
The final image removal and image subject-move counts for the examined datasets are listed in \autoref{tab:dataset-deduplicated}.
These counts correspond to the publicly available deduplication lists\footnote{\resulturl{}}.

\begin{table}[h]
\centering
\caption{\label{tab:dataset-deduplicated} The number of images that the preservative deduplication (\autoref{sec:deduplication}) removed from the dataset, and the number of images that were moved to a new subject (\autoref{sec:inter-subject-deduplication}). Note that the latter can coincide with the removal of duplicates already present in the move's target subject.}
\begin{tabular}{|r|r|r|}
\hline
      Dataset & Removed &  Moved \\
\hline
          LFW &       9 &      0 \\
     TinyFace &     354 &      0 \\
      Adience &     912 &      0 \\
CASIA-WebFace &   5,032 &     37 \\
   C-MS-Celeb & 531,018 & 13,175 \\
\hline
\end{tabular}
\end{table}

\subsection{Exact Intra-subject Deduplication}
\label{sec:exact-intra-deduplication}

The deduplication of duplicate sets that consist of exactly identical images all within the same subject is trivial,
as any one of the identical images could be randomly selected.
For the sake of reproducibility, we more specifically sort the images in ascending lexical file path order and select the first one to be kept as the deduplicated image.

The deduplication of exact inter-subject duplicate sets is less straightforward, as there are multiple different subject candidates to which the deduplicated image could be assigned to.
This is thus addressed separately in \autoref{sec:inter-subject-deduplication}.

\subsection{False Positive Correction}
\label{sec:false-positive-correction}

The image hashes used for near duplicate detection as introduced in \autoref{sec:near-duplicates} can lead to false positives,
and these false positive duplicate sets can even consist of images that a manual inspection can easily identify as different faces.
To avoid the undesired exclusion of these false positive images from the cleaned datasets,
an additional false positive correction based on face recognition is applied.

The complete near duplicate detection thus works as follows:
Near duplicate sets are first found separately by the image hash functions.
As different image hash functions can yield overlapping duplicate sets,
the next step is to merge any overlapping sets,
the result being disjunct near duplicate sets.

Then the false positive correction is carried out on each near duplicate set.
This correction step forms all comparison pairs between the images within a duplicate set,
and then filters out any images of pairs for which the similarity score is below a set threshold.

This work employs the publicly available\footnote{\url{https://github.com/IrvingMeng/MagFace}} MagFace \cite{Meng-FRwithFQA-MagFace-CVPR-2021} model with iResNet100 backbone
trained on MS1MV2 (another variant of MS-Celeb-1M \cite{Guo-Face-MSCeleb1M-ECCV-2016})
for face recognition (similarity scores), as well as for quality assessment in the following parts, as this model can be employed for both use cases.
The similarity score threshold for the false positive correction is set to $0.40$ based on the manual inspection of a randomly selected duplicate candidate subset.

\subsection{Quality-based Deduplication}
\label{sec:quality-based-deduplication}

After the false positive correction step described in \autoref{sec:false-positive-correction},
the images in each near duplicate set are sorted in descending quality score order.
The first image is then selected as the deduplicated image (\ie{} the one with the highest quality score).

These quality scores are obtained using the same MagFace model as in \autoref{sec:false-positive-correction}.
A negative infinity quality score is used for any image for which no quality score could be computed, meaning that all images with computed scores are preferred.
The ascending lexical file path order serves as a tie breaker if there are no images with computed quality scores, or if the best two quality scores are identical.

For intra-subject deduplication the selection of the deduplicated image concludes the preservative deduplication.
For inter-subject deduplication see the following \autoref{sec:inter-subject-deduplication}, as additional work is required to decide to which subject the image should be assigned to (if any).

\subsection{Comparison-based Inter-subject Deduplication}
\label{sec:inter-subject-deduplication}

Inter-subject deduplication is less straightforward than intra-subject deduplication,
as there are multiple possible subjects to which the deduplicated image could be assigned to per duplicate set.
To resolve this issue, the deduplicated image of a duplicate set is compared against the non-duplicate images of all subjects that are involved in the duplicate set.

Here ``non-duplicate images'' refers to those images that were never assigned to any duplicate set by either the exact or near duplicate detection.
If no such images remain for a candidate subject, or if no face recognition feature extraction could be applied to any of these images, then the subject is excluded from consideration.
If no candidate subjects remain, then the deduplicated image is excluded from the dataset.

The comparison of the deduplicated image to each candidate subject's non-duplicate image set is again using the same MagFace model as in \autoref{sec:false-positive-correction}.
For each subject the mean similarity score is computed across the comparisons to the deduplicated image.
The subject with the highest mean similarity score is selected as the potential target subject for the deduplicated image.

However, if the mean similarity score of the potential target subject is below a set threshold, the deduplicated image is instead excluded from the dataset.
This is done to avoid the assignment of the deduplicated image to an incorrect subject, even if that subject is the best candidate among the duplicate set's subjects.
For this the same $0.40$ similarity score threshold as in the false positive correction in \autoref{sec:false-positive-correction} is used.

Additionally, if the absolute difference between the mean similarity score of the potential target subject and the second best candidate subject is below another set threshold,
the deduplicated image is also excluded from the dataset,
to avoid uncertain subject assignments.
For this we select the similarity score threshold $0.20$ based on manual sample observations and based on the statistics of the computed mean similarity scores.

\vspace{-1.5em}
\section{\customsectiontext{Effects on Facial Biometrics}}
\label{sec:effects}

This section investigates how deduplication alters the results for face recognition and face image quality assessment experiments.
Two forms of deduplication are considered:
One is the complete removal of all images that are involved in any duplicate set according to the approach described in \autoref{sec:duplicate-detection},
and the other is the preservative deduplication described in \autoref{sec:deduplication}.

The used face recognition / quality assessment models rely on the same face image preprocessing as described in \autoref{sec:preprocessed-duplicates} for the preprocessed image duplicate detection step.
Face images for which the landmark detection failed are excluded from the experiments in this section.

\subsection{Face Recognition}
\label{sec:effect-face-recognition}

\begin{table*}[h]
\centering
\caption{\label{tab:fr-performance}
Face recognition performance in terms of the Equal Error Rate (EER) and the False Non-Match Rate (FNMR) at fixed approximated False Match Rate (FMR) values.
The ``Original'' variant refers to the unmodified datasets, the ``Full'' variant to the removal of all images involved in duplicate sets as per \autoref{sec:duplicate-detection}, and the ``Preservative'' variant to the image removal and subject-reassignment as per \autoref{sec:deduplication}.
}
\begin{tabular}{|r|r|r|r|r|}
\hline
 Dataset &      Variant &     EER &    FNMR @ FMR $\approx 1e-3$ &    FNMR @ FMR $\approx 1e-2$ \\
\hline
TinyFace &     Original & 9.5783\% & 30.0564\% & 18.8115\% \\
TinyFace &         Full & 9.7627\% & 31.1831\% & 19.8787\% \\
TinyFace & Preservative & 9.7489\% & 31.4489\% & 19.6862\% \\
\hline
Adience &     Original & 1.7576\% & 3.0122\% & 2.0063\% \\
Adience &         Full & 1.7527\% & 2.9770\% & 2.0120\% \\
Adience & Preservative & 1.7539\% & 3.0017\% & 2.0172\% \\
\hline
CASIA-WebFace &     Original & 7.1955\% & 10.2724\% & 8.7041\% \\
CASIA-WebFace &         Full & 7.2278\% & 10.3316\% & 8.7493\% \\
CASIA-WebFace & Preservative & 7.2100\% & 10.3130\% & 8.7240\% \\
\hline
C-MS-Celeb &     Original & 7.0506\% & 12.8246\% & 10.0817\% \\
C-MS-Celeb &         Full & 6.9599\% & 12.6121\% &  9.9401\% \\
C-MS-Celeb & Preservative & 6.9094\% & 15.4464\% & 10.0007\% \\
\hline
\end{tabular}
\end{table*}

\begin{table}[h]
\centering
\caption{\label{tab:edc-magface} FNMR and FMR EDC pAUC values for the $[0\%, 20\%]$ comparison discard fraction range, using MagFace for quality assessment.
}
\begin{tabular}{|r|r|r|r|}
\hline
      Dataset &  Variant &    FNMR &     FMR \\
\hline
     TinyFace & Original & 0.859\% & 0.977\% \\
     TinyFace &     Full & 0.864\% & 0.972\% \\
     TinyFace & Preserv. & 0.865\% & 0.973\% \\
\hline
      Adience & Original & 0.708\% & 0.992\% \\
      Adience &     Full & 0.731\% & 0.994\% \\
      Adience & Preserv. & 0.726\% & 0.996\% \\
\hline
CASIA-WebFace & Original & 0.820\% & 1.002\% \\
CASIA-WebFace &     Full & 0.820\% & 1.002\% \\
CASIA-WebFace & Preserv. & 0.820\% & 1.002\% \\
\hline
   C-MS-Celeb & Original & 0.814\% & 1.000\% \\
   C-MS-Celeb &     Full & 0.804\% & 1.000\% \\
   C-MS-Celeb & Preserv. & 0.802\% & 1.002\% \\
\hline
\end{tabular}
\end{table}

\begin{table}[h]
\centering
\caption{\label{tab:edc-crfiqal} FNMR and FMR EDC pAUC values for the $[0\%, 20\%]$ comparison discard fraction range, using CR-FIQA(L) for quality assessment.
}
\begin{tabular}{|r|r|r|r|}
\hline
      Dataset &  Variant &    FNMR &     FMR \\
\hline
     TinyFace & Original & 0.801\% & 0.890\% \\
     TinyFace &     Full & 0.803\% & 0.890\% \\
     TinyFace & Preserv. & 0.809\% & 0.885\% \\
\hline
      Adience & Original & 0.865\% & 0.958\% \\
      Adience &     Full & 0.880\% & 0.959\% \\
      Adience & Preserv. & 0.875\% & 0.957\% \\
\hline
CASIA-WebFace & Original & 0.799\% & 1.000\% \\
CASIA-WebFace &     Full & 0.800\% & 0.999\% \\
CASIA-WebFace & Preserv. & 0.799\% & 0.999\% \\
\hline
   C-MS-Celeb & Original & 0.812\% & 0.997\% \\
   C-MS-Celeb &     Full & 0.800\% & 0.997\% \\
   C-MS-Celeb & Preserv. & 0.798\% & 0.999\% \\
\hline
\end{tabular}
\end{table}

The following face recognition experiments require the selection of mated and non-mated \cite{Vocabulary} pairs of face images.

For the mated pairs, one approach could be to select all possible pairs for each subject.
This may however result in a substantially increased number of mated pairs for subjects with comparatively larger image counts within a dataset,
since a subject with $N$ images would yield $(N\cdot (N-1))/2$ mated pairs,
and each image would consequently be involved in $N-1$ mated pairs.

\begin{figure}[!h]
\centering
\begin{minipage}[t]{0.2\textwidth}
\centering
All mated pairs\\ per subject:
\includegraphics[width=0.8\textwidth,valign=t]{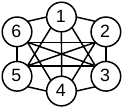}
\end{minipage}
\begin{minipage}[t]{0.2\textwidth}
\centering
``Circular'' mated pairs per subject:
\includegraphics[width=0.8\textwidth,valign=t]{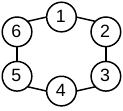}
\end{minipage}
\caption{\label{fig:mated-pair-selection}Two mated pair selection approaches. The numbered graph nodes represent the ordered face images of a single subject, with the graph edges representing the selected mated pairs.}
\end{figure}

But since these experiments are supposed to compare results for the datasets with and without duplicates,
a balanced number of mated pairs per image is preferred.
Mated pairs are thus instead selected in a ``circular'' manner for each subject in a dataset,
as illustrated in \autoref{fig:mated-pair-selection}.
This simply means that every image with index $i$ forms a mated pair with the next image with index $i+1$.
The last image also forms a pair with the first image if the subject has more than two images.
A subject with more than two images thus results in a number of mated pairs equal to the number of images,
and a subject with exactly two images results in a single mated pair.
Each image is consequently always involved in exactly two mated pairs for subjects with more than two images,
or one mated pair for subjects with two images.
The lexicographically ascending order of the image paths is used as the image order.
An additional benefit of this  ``circular'' mated pair selection is the reduction of the required computational resources for the experiments,
since the number of all possible mated pairs would be substantially higher (over 300 million in total).

The following lists the mated pair count per dataset, as well as the number of subjects implicitly excluded due to only containing a single image:

\begin{itemize}
\item TinyFace: 11,881 (153 excluded subjects).
\item Adience: 18,093 (815 excluded subjects).
\item CASIA-WebFace: 494,284 (0 excluded subjects).
\item C-MS-Celeb: 6,457,562 (123 excluded subjects).
\end{itemize}

A number of non-mated pairs equal to the number of mated pairs is randomly selected per dataset.

The MagFace model previously described in \autoref{sec:false-positive-correction} is employed for face recognition.
The similarity scores are then used to assess the face recognition performance in terms of the False Non-Match Rate (FNMR), the False Match Rate (FMR), and the Equal Error Rate (EER) \cite{Vocabulary}.
\autoref{tab:fr-performance} shows the results.
There are mostly minor differences between the variants with and without duplicates,
and duplicate removal both increased and decreased error rates in different cases.

\subsection{Quality Assessment}
\label{sec:effect-quality-assessment}

The effect of the duplicate removal on face image quality assessment is examined in terms of the change of partial Area Under Curve (pAUC) values for Error versus Discard Characteristic (EDC) curves \cite{Schlett-EDC-considerations-2023,Grother-SampleQualityMetricERC-PAMI-2007},
using the FNMR and FMR as the EDC errors.
This evaluation is based both on quality scores and the previously computed similarity scores.
The quality scores are computed by two state-of-the-art models,
which aim to assess the face image quality in terms of the biometric utility for face recognition \cite{Vocabulary}.
The first model is again the MagFace model described in \autoref{sec:false-positive-correction},
and the second model is CR-FIQA(L) \cite{Boutros-CR-FIQA-CVPR-2023},
which is likewise publicly available\footnote{\url{https://github.com/fdbtrs/CR-FIQA}}.
\autoref{tab:edc-magface} and \autoref{tab:edc-crfiqal} show the results using MagFace and CR-FIQA(L), respectively.
The pAUC values are computed for the $[0\%, 20\%]$ discard fraction range instead of the full EDC curves,
since higher discard fractions are not usually considered as operationally relevant \cite{Schlett-EDC-considerations-2023}.
Similarly to the face recognition results of the prior subsection,
there are minor differences which include both increases and decreases in the error values.

\section{\customsectiontext{Summary}}
\label{sec:summary}

The presented exact and near face image duplicate detection approach based on file hashes and image hashes (\autoref{sec:duplicate-detection}), including the additional use of preprocessed face images (\autoref{sec:preprocessed-duplicates}), found duplicates in all five examined web-scraped datasets (\autoref{sec:datasets}).
With the exception of the LFW dataset, over 1\% of all images in each original dataset are duplicate candidates, which ranges from hundreds to hundreds of thousands of duplicate candidates in absolute numbers (\autoref{sec:datasets}).
While most duplicate set images belong to a single dataset subject (intra-subject duplicates), there also are some that belong to multiple (inter-subject duplicates), especially in the C-MS-Celeb dataset (\autoref{sec:datasets}).

Preservative deduplication steps (\autoref{sec:deduplication}) were applied after the initial duplicate detection.
This comprised the mitigation of false positive duplicate detection,
the selection of the highest quality face images per duplicate set (according to a face image quality assessment model),
and the assignment of deduplicated face images to the most fitting subject within inter-subject duplicate sets (or none if uncertain).
The final deduplication data for the examined datasets is publicly available\footnote{\resulturl{}}.

Minor effects due to duplicate removal were observed on face recognition and face image quality assessment results in the experiments (\autoref{sec:effects}).

Finding duplicates in all examined datasets does indicate that such accidental inclusion of duplicates could be a common occurrence for web-scraped face image datasets,
so that any potential future dataset construction of this kind should consider implementing a duplicate filter.
It may further be sensible to examine other existing web-scraped datasets, as they could likewise contain duplicates.

\section*{\uppercase{Acknowledgements}}

This research work has been funded by the German Federal Ministry of Education and Research and the Hessian Ministry of Higher Education, Research, Science and the Arts within their joint support of the National Research Center for Applied Cybersecurity ATHENE.
This project has received funding from the European Union’s Horizon 2020 research and innovation programme under grant agreement No 883356.
This text reflects only the author’s views and the Commission is not liable for any use that may be made of the information contained therein.

\bibliographystyle{apalike}
{\small
\bibliography{bib/references}}

\end{document}